\documentclass[runningheads]{llncs}
\usepackage{xcolor}
\usepackage{times}
\usepackage{soul}
\usepackage{url}
\usepackage[hidelinks]{hyperref}
\usepackage[utf8]{inputenc}
\usepackage[small]{caption}
\usepackage{graphicx}
\usepackage{amsmath}
\usepackage{booktabs}
\usepackage{algorithm}
\usepackage{algorithmic}
\usepackage{subcaption}
\usepackage{multirow}
\usepackage{subcaption}
\usepackage{float}
\usepackage{xspace}
\usepackage{listings}

\usepackage[nameinlink]{cleveref}

\newcommand{\essence}[0]{\textsc{Essence}\xspace}
\newcommand{\conjure}[0]{\textsc{Conjure}\xspace}
\newcommand{\eprime}[0]{\textsc{Essence Prime}\xspace}
\newcommand{\savilerow}[0]{\textsc{Savile Row}\xspace}
\newcommand{\minion}[0]{\textsc{Minion}\xspace}

\usepackage{tikz}
\usetikzlibrary{shapes.geometric, arrows, positioning}
\tikzstyle{startstop} = [rectangle, rounded corners, minimum width=2cm, minimum height=1cm,text centered, draw=black]
\tikzstyle{io} = [trapezium, trapezium left angle=70, trapezium right angle=110, minimum width=2cm, minimum height=1cm, text centered, draw=black]
\tikzstyle{process} = [rectangle, minimum width=35mm, minimum height=1cm, text centered, text width=38mm, draw=black]
\tikzstyle{decision} = [diamond, minimum width=2cm, minimum height=1cm, text centered, draw=black]
\tikzstyle{arrow} = [thick,->,>=stealth]

\begin{document}

\title{Towards Improving Solution Dominance with Incomparability Conditions: \\ A case-study using Generator Itemset Mining}
\titlerunning{Improving Solution Dominance with Incomparability Conditions}

\author{Gökberk Koçak\inst{1}, Özgür Akgün\inst{1}, Tias Guns\inst{2}, Ian Miguel\inst{1}}
\authorrunning{G Koçak et al.}

\institute{
School of Computer Science, University of St Andrews, UK\\
\email{\{gk34,ozgur.akgun,ijm\}@st-andrews.ac.uk}\and
Vrije Universiteit Brussel\\
\email{Tias.Guns@vub.be}
}

\maketitle

\begin{abstract}

Finding interesting patterns is a challenging task in data mining. Constraint based mining is a well-known approach to this, and one for which constraint programming has been shown to be a well-suited and generic framework.
Dominance programming has been proposed as an extension that can capture an even wider class of constraint-based mining problems, by allowing to compare relations between patterns.
In this paper, in addition to specifying a dominance relation, we introduce the ability to specify an incomparability condition. Using these two concepts we devise a generic framework that can do a batch-wise search that avoids checking incomparable solutions. We extend the \essence{} language and underlying modelling pipeline to support this.
We use generator itemset mining problem as a test case and give a declarative specification for that.
We also present preliminary experimental results on this specific problem class with 
a CP solver backend to
show that using the incomparability condition during search can improve the efficiency of dominance programming and reduces the need for post-processing to filter dominated solutions.


\end{abstract}
\section{Introduction}



Pattern Mining is the process of finding interesting patterns in large data-sets. Common pattern mining tasks include problems like the well known problem of frequent itemset mining (FIM) (sets of items that occur together frequently) from transactional databases. Standard pattern mining tasks that require enumerating all frequent itemsets are best performed using specialised tools and algorithms~\cite{agrawal1994fast,zaki2000scalable,han2004mining}. However, a complete enumeration of all frequent itemsets is rarely what a practitioner needs, since the number of all frequent itemsets can be very large. The main goal of pattern mining is to find a smaller number of interesting patterns for further analysis. Domain-specific side constraints~\cite{bonchi2004closed} and methods for compactly representing the outcome of a particular pattern mining task~\cite{pasquier1999discovering,soulet2014efficiently,szathmary2007towards} have been proposed to increase the utility of constraint-based pattern mining. While these methods allow us to focus on interesting patterns, and represent solution sets compactly, they also result in a significantly more difficult data mining task.

Constraint Programming (CP) is a general purpose method for specifying decision problems in a declarative language and finding solutions to these problems using highly efficient black-box solvers. Recent work demonstrates the utility of CP for performing constraint-based data mining tasks~\cite{de2008constraint,guns2013miningzinc,guns2017miningzinc}. The main advantage of these methods is their generic nature and hence flexibility: once a CP model is developed for a certain pattern mining task this model can be extended with additional side constraints easily. This contrasts with specialised algorithms, where incorporating domain knowledge is often difficult, side constraints often help a black-box constraint solver. An example of a generic CP-based language and framework for pattern mining is MiningZinc~\cite{guns2013miningzinc}, which was built on top of MiniZinc to allow the easy specification of data mining problems in a CP modelling environment.

While many constraints can be expressed as standard constraints, a number of tasks do not allow the addition of arbitrary side constraints. The most well-known example is closed frequent itemset mining with side constraints that are not monotone (for example a maximum size constraint)~\cite{bonchi2004closed}.
Enforcing the general property of closedness/maximality for frequent itemset mining can be solved by adding constraints among solutions. In a closed frequent itemset mining (CFIM) task, a frequent itemset is only a solution if its support is greater than all of its supersets. Constraint Dominance Programming (CDP) has been suggested as a way of formulating such properties in a general way~\cite{negrevergne2013dominance,guns2018solution}. A CDP model specifies, in addition to the main model where the decision variables and constraints relating to a single solution are declared in the usual way, constraints among solutions using dominance-nogoods. The operational semantics of dominance-nogoods corresponds to adding a new blocking constraint after each solution. This way, potential solutions that are dominated by a previously found solution are blocked. Following this semantics, CDP always finds all non-dominated solutions; however, without a perfect search ordering it is likely to find a number of dominated solutions as well. Dominated solutions can then be removed using a post-processing step~\cite{negrevergne2013dominance}.

In this paper, we add the necessary language and search capabilities described by Guns et al.~\cite{guns2018solution} to \essence. Furthermore, we propose an extension to CDP with an incomparability condition between solutions. In addition to specifying the dominance relation between solutions, we also specify the condition which makes the solutions incomparable. Hence, in an enumaration task using the incomparability condition system can enumarate all the solutions that are incomparable to each other in one solver call and add the blocking no-goods later. This abstracts the idea of level-wise search for common pattern mining~\cite{chan2003mining,bonchi2003examiner} in CP. Level-wise search in pattern mining using CP was previously explored in \cite{koccak2018closed}. Using dominance programming logic demonstrates the abstraction of level-wise search more clearly and supports the application of incomparability as an extension to dominance programming without enforcing level-wise search.

\textbf{Contributions.} We extend \essence{} to support CDP. We define and implement \textit{incomparability} as an enhancement to pure-CDP (our version referred to as CDP+I in the rest of the paper). It addresses the main bottleneck of dominance programming: CDP+I often generates many fewer blocking constraints, it allows the natural specification of the search order, and it can eliminate the post-processing step that is often required to filter dominated solutions.

\section{Constraint Dominance Programming}

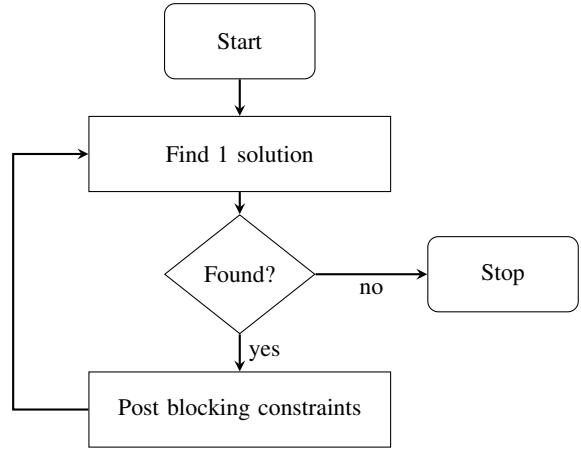
\begin{figure}[t]
    \centering
    \begin{tikzpicture}[node distance=1.5cm]
        \node (start) [startstop] {Start};
        \node (run2) [process, below of=start] {Find 1 solution};
        \node (sol) [decision, below of=run2,yshift=-1mm] {Found?};
        \node (encode2) [process, below of=sol, yshift=-3mm] {Post blocking constraints};
        \node (stop) [startstop, right of=sol, xshift=2cm] {Stop};
        \draw [arrow] (start) -- (run2);
        \draw [arrow] (sol) -- node[anchor=north] {no} (stop.west);
        \draw [arrow] (sol) -- node[anchor=west] {yes} (encode2);
        \draw [arrow] (encode2.west) -- +(-1,0) --  +(-1,3.4)  -- (run2.west);
        \draw [arrow] (run2) -- (sol);
    \end{tikzpicture}
    \caption{Flowchart for Constraint Dominance Programming (CDP)}
    \label{fig:sd_flow}
\end{figure}

A constraint dominance problem is a constraint satisfaction problem extended with dominance nogoods~\cite{guns2018solution}. A dominance nogood is a blocking constraint that can be generated from an existing solution, and is used to prune all solutions dominated by the solution at hand.

When solving a constraint dominance problem, the goal is to enumerate all non-dominated solutions. Operationally this is achieved by finding a solution $S$, posting blocking constraints to disallow solutions dominated by $S$, and using the modified model to find the next solution. This process creates as many dominance blocking constraints as there are solutions. Moreover, without a perfect search order, it may produce dominated solutions in addition to all the non-dominated solutions. A post-processing step is needed to remove some dominated solutions which can be found in the intermediate solver calls. Thus, it requires calling the low-level solver at least $n+1$ times for $n$ solutions. \Cref{fig:sd_flow} gives a flowchart for CDP.

To understand the dominance problem a simple example can be given. If we look at~\Cref{eq:dom}, the dominance condition indicates that the future possible solutions shouldn't be a subset of any found solution. After finding one of the solutions, let's say $\{1,2\}$, $\{1\}, \{2\}$ are blocked to be found later on. Respectively, $\{1,3\}$ will block only $\{3\}$ since $\{1\}$ is already blocked. 

\begin{equation}
    \begin{Bmatrix}
\text{Example Solutions} = \{1,2\}, \{1,3\}
\\ \text{Dominance Condition} = \text{ } \not\exists S_{future} \text{ } \forall S \text{ } | \text{ } S_{future} \subseteq S 
\\ \text{Dominated Solutions} = \{1\}, \{2\}, \{3\}
\end{Bmatrix}
\label{eq:dom}
\end{equation}

Pure CDP has a number of shortcomings in the context of constraint based mining.
    
\begin{itemize}
    \item When the number of solutions is large, CDP's requirement of calling the solvers once for solution creates an overhead, as it incrementally adds a new dominance blocking constraint after each solution. This creates an unnecessary overhead, and in addition it reduces the utility of learned clauses in a learning solver. It might also evaluate the same sub-trees of searches multiple times, since this information is lost at the end of solver call.
    \item Without a good search ordering, CDP might enumerate dominated solutions as well. The number of dominated solutions is typically orders of magnitude greater than of non-dominated solutions.
    \item Post-processing is required to remove dominated solutions found during search.
\end{itemize}


We observe that a significant number of pairs of solutions $A$ and $B$ in the solution set of a CDP tend to be incomparable to each other. Two solutions $A$ and $B$ are incomparable if neither $A$ dominates be, nor $B$ dominates $A$. For these pairs of solutions, the dominance blocking constraint generated from either solution is irrelevant when searching for the other solution. In the next section we present an explicit way of capturing such conditions declaratively in a new incomparability function statement.

%

\subsection{Incomparability in Constraint Dominance Programming}

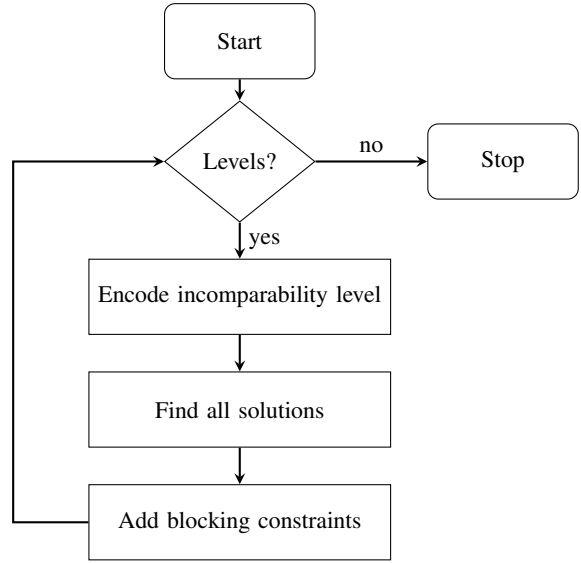
\begin{figure}[t]
    \centering
    \begin{tikzpicture}[node distance=1.5cm]
        \node (start) [startstop] {Start};
        \node (levels) [decision, below of=start,yshift=-1mm] {Levels? };
        \node (encodei) [process, below of=levels, yshift=-3mm] {Encode incomparability level};
        \node (run) [process, below of=encodei] {Find all solutions};
        \node (encoded) [process, below of=run] {Add blocking constraints};
        \node (stop) [startstop, right of=levels, xshift=2cm] {Stop};
        \draw [arrow] (start) -- (levels);
        \draw [arrow] (levels) -- node[anchor=west] {yes} (encodei);
        \draw [arrow] (levels) -- node[anchor=south] {no} (stop);
        \draw [arrow] (encodei) -- (run);
        \draw [arrow] (run) -- (encoded);
        \draw [arrow] (encoded.west) -- +(-1,0) -- +(-1,4.8)  -- (levels.west);
    \end{tikzpicture}
    \caption{Flowchart for CDP enhanced with Incomparability (CDP+I)}
    \label{fig:sr_flow}
\end{figure}

We define a new type of statement to specify incomparable solutions explicitly. This statement is only allowed in a CDP problem specification, i.e. a problem specification which contains a dominance relation statement.
The dominance relation statement defines the dominance relation itself, similarly to the dominance nogoods introduced in~\cite{guns2018solution}.
The incomparability function statement provides a function ($I$) mapping any solution to a single value that has an ordered \essence{} type (typically an integer).
Two solutions $A$ and $B$ with $I(A) = I(B)$ are said to be incomparable, i.e. neither $A$ dominates $B$, nor $B$ dominates $A$.
The addition to the previous example we looked at~\Cref{eq:dom} would be $|S|$. Any found solution set $S_1$ do not dominate any other solution $S_2$ with the same cardinality ($|S_1|=|S_2$) within the given subset relation. Same situation applies the other way around; $S_2$ do not dominate $S_1$. 
The incomparability function partitions the search space. Since we enumerate all solutions for each part of the partition, it is necessarily sound in not losing non-dominated solutions. 

Operationally we make use of this explicit incomparability statement by enumerating all solutions that have an equal incomparability function value. This avoids the need to add any blocking constraints after each solution that has the same incomparability value. Then, all of the necessary blocking constraints are added at once before moving to the next incomparability level. This reduces the number of solver calls required, reduces the total number of dominance blocking constraints maintained, and allows the use of efficient solution-enumeration solvers. In addition, thanks to the explicit search order specified in the incomparability function statement the proposed system will produce fewer (or in the best case none) dominated solutions. CDP enhanced with an explicit incomparability statement (CDP+I) is implemented in \conjure{}~\cite{akgun2013automated,akgun2014breaking,akgun2011extensible} and \savilerow{}~\cite{nightingale2017automatically}; \Cref{fig:sr_flow} gives a flowchart for CDP+I. 



\section{Architecture in \conjure{} and \savilerow{}}

\essence{} is a high-level constraint modelling languages that natively supports set, multi-set, sequence, function, relation domain constructors and domains that are made of these domain constructors in an arbitrarily nested manner~\cite{frisch2008ssence}. The pattern mining models listed in this paper make use of set, sequence, and multi-set domains to represent the frequent itemset decision variables and the transaction databases themselves. Operators such as \texttt{subsetEq} and quantification over decision variables with set domains are used to specify the models succinctly and abstractly. \conjure{} is then used to generate concrete constraint programming models in the lower-level \eprime{} language. During this step, \conjure{} can generate several alternative models since there are multiple representation options for set, multi-set and sequence decision variables and parameters. In this paper we do not compare different reformulations, instead we rely on the built-in heuristics of \conjure{} to produce a reasonable model.

\eprime{} is a solver-independent constraint modelling language that is comparable in terms of features to the MiniZinc language~\cite{nethercote2007minizinc}. It offers Boolean, integer and matrix decision variables, a rich collection of operators that work on these types, and a number of global constraints. \savilerow{} is able to target several backend solvers. We experiment with a CP backend (\minion{}) and a SAT backend of \savilerow{} in this paper. We extend \eprime{} to represent \savilerow{} to handle both dominance relation and incomparability function statements.

\savilerow{} is an optimising constraint modelling assistant~\cite{nightingale2017automatically}. Among others, it performs common subexpression elimination (CSE) and dynamic probing for domain pruning. These modelling optimisations help all configurations of our computational experiments presented in this paper. Specifically for the domain pruning optimisation \savilerow{} calls \minion{} in a special mode to only perform singleton arc consistency on the bounds of the variables (SACBounds~\cite{nightingale2014automatically}) and return potentailly reduced domains. This optimisation allows us to reduce the number of incomparability levels completely automatically. During SACBounds reasoning, if there are no frequent itemsets with cardinality above a certain value, the corresponding domain is pruned accordingly. This can result in entire incomparability levels not having to be instantiated.


When targeting \minion{}, \savilerow{} is used to produce an input file using the standard pipeline. Then, a collection of constraints are added to enforce a level as deduced from the incomparability function statement. These constraints are added in a way that would allow their easy removal from the \minion{} input file. Once all solutions are enumerated for a certain incomparability level; the previous level constraints are removed, all dominance blocking constraints are added at once, and the next level constraints are added. The dominance blocking constraints are added by substituting the solution values inside the dominance relation expression specified in the model. This process continues until there are no more incomparability levels left.




%

For implementing CDP+I, \conjure{} is modified to handle and translate dominance relation and incomparability function statements. The heavy lifting is done by \savilerow{} - it recognises the existence of these new statements and follows the flowchart given in \Cref{fig:sd_flow} or \Cref{fig:sr_flow}, depending on whether an incomparability function statement exists or not. In these cases, respectively, it calls the backend solver (for one solution) repeatedly and adds blocking constraints after each solution, or it calls the backend solver (for all solutions) and adds blocking constraints after each level.

\section{CDP+I Model Example for Itemset Mining Rroblem\label{sec:models}}

We present a CDP+I model  for the generator itemset mining problem\footnote{We make our \essence{} model and experimental results available in a Github repository \url{https://github.com/stacs-cp/ModRef2019-Dominance}.}.

This model operates on a transactional dataset in the form of a multi-set of set of items. The decision variables are: we always try to find a set of items to represent the pattern (decision variable name is \texttt{itemset}) and its support using integers (decision variable name is \texttt{support}).

We use two side constraints on the pattern for each model; minimum value~\cite{guns2013miningzinc} and maximum cost~\cite{bonchi2004closed,bonchi2007extending}. 
Moreover, the minimum value constraint is monotone and the maximum cost constraint is not monotone; hence we use these two constraints to demonstrate the correct handling of any side constraints independent of whether they are monotone or not. The corresponding \essence{} specification is:

\begin{lstlisting}[numbers=none]
such that
  (sum item in itemset . values[item]) >= min_value,
  (sum item in itemset . costs[item]) <= max_cost
\end{lstlisting}

\subsection{Generator Itemset Mining}

Frequent itemset mining is a standard data mining problem where the task is enumerating itemsets whose \textit{support} is above a given threshold value. Support is the number of transaction that contain the pattern itemset as a subset.

Generator itemsets (also called free itemsets or key itemsets~\cite{boulicaut2000approximation,boulicaut2001mining}) are a related compressed representation of the all frequent itemsets. A generator itemset is a frequent itemset which does not have any frequent \textit{sub}sets with the same support. Thus, They represent the minimal frequent itemsets and every other frequent itemset can be built from generators in a small to larger manner. 

Generator itemsets are useful as part of a larger association rule mining task, together with closed frequent itemsets to find minimal non-redundant association rules~\cite{kryszkiewicz1998representative}.

To express them in the dominance logic, we can try to find minimal frequent itemsets which generates the bigger ones. The dominance will indicate smaller itemsets to dominate the bigger ones with same support`.

\begin{lstlisting}[numbers=none]
dominance_relation (fromSolution(itemset) subsetEq itemset)
                    -> (support != fromSolution(support))
\end{lstlisting}

The dominance relation follows the definition very closely. A frequent itemset is a generator itemset if its support is not equal to the support of any of its subsets.

\begin{lstlisting}[numbers=none]
incomparability_function ascending |itemset|
\end{lstlisting}

The incomparability function for generator itemsets is very similar to that of closed itemsets: it uses the itemset cardinality. Two sets $A$ and $B$ are guaranteed not to dominate each other if their cardinalities are the same. This condition is complete when paired with an ascending direction of search on the itemset cardinality. In contrast to CFIM, we first find smaller itemsets which are generator itemsets by definition since they do not have any frequent subsets. Then, dominance blocking constraints are added and we only find generator itemsets in the successive levels.

By intuition, If we start of from the empty set and look generators increasing on cardinality, it is sounds that the system won't find any dominated solutions.

\section{Computational experiments}



The proposed example model has a minimum value and maximum cost side constraints to demonstrate their ability to handle arbitrary side constraints. We uniformly randomly generate values between 0 and 5 for minimum value and maximum cost. In addition, we generate a threshold for the minimum value and the maximum cost constraints as well. We systematically generate several candidate instances and choose instances which have a reasonable number of solutions (in the 10,000s at most for all problem classes except minimal rare itemset mining and in the 100,000s for minimal rare) and those that can be solved within our time limit of 6-hours\footnote{We make the problem instances available in our Github repository \url{https://github.com/stacs-cp/ModRef2019-Dominance}}. We generate instances at 5 frequency levels: 10\%, 20\%, 30\%, 40\%, 50\%. For CFIM we use the instances published in the appendix of \cite{koccak2018closed}.


We employ three configurations to solve each instance. The first two do not include the incomparability condition, hence they are pure CDP. The difference between the first two configurations is that the second is tuned to have the same search order as the third option CDP+I. The first configuration (henceforth called CDP\_default\_order) uses the default variable branching order, which is the order of the appearance of the decision variables. The reason of having CDP\_default\_order is that it can be generated from the high level specifications without having expertise. A naive pattern miner without much CP experience may use the high level CP model without tweaking the search parameters. The CDP configuration (i.e CDP\_level\_order) branches on the level information (cardinality) in the first place just like CDP+I enforces in its incomparability condition. The incomparability condition can also capture the search order aspect without requring CP expertise for the user.


We run our experiments on two identical 32 core AMD Opteron 6272 machines, at 2.1 GHz and with 256GB memory. We run 31 cores in parallel on each machine and left one core idle to account for system processes. Each separate experiment was given a single CPU core, 8GB of memory and a 6-hours time limit. 

\subsection{Datasets}

We have used 12 datasets from UCI indirectly and took them from CP4IM \footnote{\url{https://dtai.cs.kuleuven.be/CP4IM/datasets/}}. The characteristics of these datasets can be seen in the figure \Cref{tab:db}.

\begin{table}[h!]
\centering
\begin{tabular}{|l|r|r|r|}
\hline 
dataset  & transaction & items & density \\ \hline 
Anneal                & 812         & 93    &  45\%  \\
Audiology             & 216         & 148   &  45\%  \\
Australian-credit                   & 653         & 128   &  41\% \\
German-credit                & 1000        & 112   &  34\%  \\
Heart-cleveland                 & 296         & 95    &  47\% \\
Hepatitis               & 137         & 68    &  50\%  \\
Hypothyroid                  & 3247        & 88    &  49\%  \\
Kr-vs-kp                & 3196        & 73    &  49\% \\
Lymph                & 148         & 68    &  40\%  \\
Primary-tumor                 & 336         & 31    &  48\%  \\
Vote                  & 435         & 48    &  33\%  \\
Zoo                   & 101         & 36    &  44\%  \\
\hline
\end{tabular}
\caption{The dataset which are used in our experiments \label{tab:db}}
\end{table}

\subsection{All results}

All results can be seen in \Cref{fig:all}. The missing results indicate a time out.

\begin{table}[h!]

    \centering
\centering
\begin{tabular}{|l|r|r|r|r|r|r|r|r|r|r|}
\hline
            & \multicolumn{1}{l|}{}        & \multicolumn{3}{c|}{CDP\_default\_order}                                             & \multicolumn{3}{c|}{CDP\_level\_order}                                               & \multicolumn{3}{c|}{CDP+I}                                                           \\ \hline
Instance    & \multicolumn{1}{l|}{Nb Sols} & \multicolumn{1}{l|}{Time} & \multicolumn{1}{l|}{Blocks} & \multicolumn{1}{l|}{Calls} & \multicolumn{1}{l|}{Time} & \multicolumn{1}{l|}{Blocks} & \multicolumn{1}{l|}{Calls} & \multicolumn{1}{l|}{Time} & \multicolumn{1}{l|}{Blocks} & \multicolumn{1}{l|}{Calls} \\ \hline
audio\_40   & 1808                         & *                         & *                           & *                          & *                         & *                           & *                          & 1234.34                   & 34677                       & 65                         \\ 
aus\_40     & 18                           & *                         & *                           & *                          & *                         & *                           & *                          & 7961.51                   & 699                         & 51                         \\ 
aus\_50     & 4488                         & *                         & *                           & *                          & *                         & *                           & *                          & 734.53                    & 190161                      & 49                         \\ 
german\_20  & 6                            & *                         & *                           & *                          & *                         & *                           & *                          & 11295.05                  & 161                         & 38                         \\ 
german\_30  & 144                          & *                         & *                           & *                          & *                         & *                           & *                          & 2105.29                   & 4214                        & 38                         \\ 
german\_40  & 2338                         & *                         & *                           & *                          & *                         & *                           & *                          & 478.61                    & 75112                       & 38                         \\ 
german\_50  & 398                          & 4729.19                   & 79401                       & 399                        & 4382.17                   & 79401                       & 399                        & 116.54                    & 9954                        & 30                         \\ 
heart\_30   & 21                           & *                         & *                           & *                          & *                         & *                           & *                          & 4529.58                   & 710                         & 45                         \\ 
heart\_40   & 4928                         & *                         & *                           & *                          & *                         & *                           & *                          & 792.99                    & 186202                      & 45                         \\ 
heart\_50   & 527                          & 3327.65                   & 139128                      & 528                        & 3324.67                   & 139128                      & 528                        & 57.88                     & 17461                       & 39                         \\ 
hepatit\_20 & 1819                         & *                         & *                           & *                          & *                         & *                           & *                          & 4892.46                   & 38811                       & 34                         \\ 
hepatit\_30 & 586                          & *                         & *                           & *                          & *                         & *                           & *                          & 487.86                    & 14458                       & 34                         \\ 
hepatit\_40 & 9231                         & *                         & *                           & *                          & *                         & *                           & *                          & 170.15                    & 266804                      & 34                         \\ 
hepatit\_50 & 3610                         & 12510.53                  & 6517855                     & 3611                       & 12059.16                  & 6517855                     & 3611                       & 31.96                     & 108546                      & 34                         \\ 
krvskp\_40  & 1230                         & *                         & *                           & *                          & *                         & *                           & *                          & 9967.5                    & 28051                       & 36                         \\ 
krvskp\_50  & 427                          & *                         & *                           & *                          & *                         & *                           & *                          & 5171.46                   & 9728                        & 35                         \\ 
lymph\_10   & 2391                         & *                         & *                           & *                          & *                         & *                           & *                          & 940.83                    & 34299                       & 27                         \\ 
lymph\_20   & 12996                        & *                         & *                           & *                          & *                         & *                           & *                          & 511.67                    & 227878                      & 27                         \\ 
lymph\_30   & 9811                         & *                         & *                           & *                          & *                         & *                           & *                          & 160                       & 192797                      & 27                         \\ 
lymph\_40   & 4181                         & *                         & *                           & *                          & *                         & *                           & *                          & 43.38                     & 89209                       & 27                         \\ 
lymph\_50   & 1044                         & 563.32                    & 545490                      & 1045                       & 570.84                    & 545490                      & 1045                       & 8.16                      & 20259                       & 23                         \\ 
tumor\_20   & 1998                         & 4621.02                   & 1997001                     & 1999                       & 3416.57                   & 1997001                     & 1999                       & 6.24                      & 19339                       & 15                         \\ 
tumor\_30   & 1191                         & 1326.16                   & 709836                      & 1192                       & 978.22                    & 709836                      & 1192                       & 3.86                      & 11244                       & 15                         \\ 
tumor\_40   & 310                          & 122.04                    & 48205                       & 311                        & 94.63                     & 48205                       & 311                        & 2.05                      & 3021                        & 15                         \\ 
tumor\_50   & 61                           & 18.47                     & 1891                        & 62                         & 15.3                      & 1891                        & 62                         & 1.62                      & 554                         & 14                         \\ 
vote\_10    & 9962                         & *                         & *                           & *                          & *                         & *                           & *                          & 76.28                     & 109203                      & 16                         \\ 
vote\_20    & 2983                         & 16142.82                  & 4450636                     & 2984                       & *                         & *                           & *                          & 21.27                     & 33929                       & 16                         \\ 
vote\_30    & 268                          & 365.31                    & 36046                       & 269                        & 268.43                    & 36046                       & 269                        & 8.49                      & 3243                        & 16                         \\ 
vote\_40    & 5                            & 4.65                      & 15                          & 6                          & 3.08                      & 15                          & 6                          & 4.9                       & 63                          & 16                         \\ 
zoo\_20     & 4286                         & *                         & *                           & *                          & *                         & *                           & *                          & 24.32                     & 45002                       & 16                         \\ 
zoo\_30     & 1132                         & 740.21                    & 641278                      & 1133                       & 565.09                    & 641278                      & 1133                       & 2.27                      & 11959                       & 16                         \\ 
zoo\_40     & 64                           & 11.95                     & 2080                        & 65                         & 10.38                     & 2080                        & 65                         & 0.62                      & 732                         & 16                         \\ 
zoo\_50     & 7                            & 1.29                      & 28                          & 8                          & 1.06                      & 28                          & 8                          & 0.45                      & 67                          & 14                         \\ \hline
\end{tabular}

\caption{All results; while Sols and Calls columns are self explanatory and indicate the number of solutions and number of solver calls respectively, Blocks column means the number of blocking constraints used in all solver calls. * indicates experiment timed out.
    \label{fig:all} }    
\end{table}
\subsubsection{On Number of Solutions}

\begin{figure}[t]
    \centering
    \includegraphics[width=\textwidth]{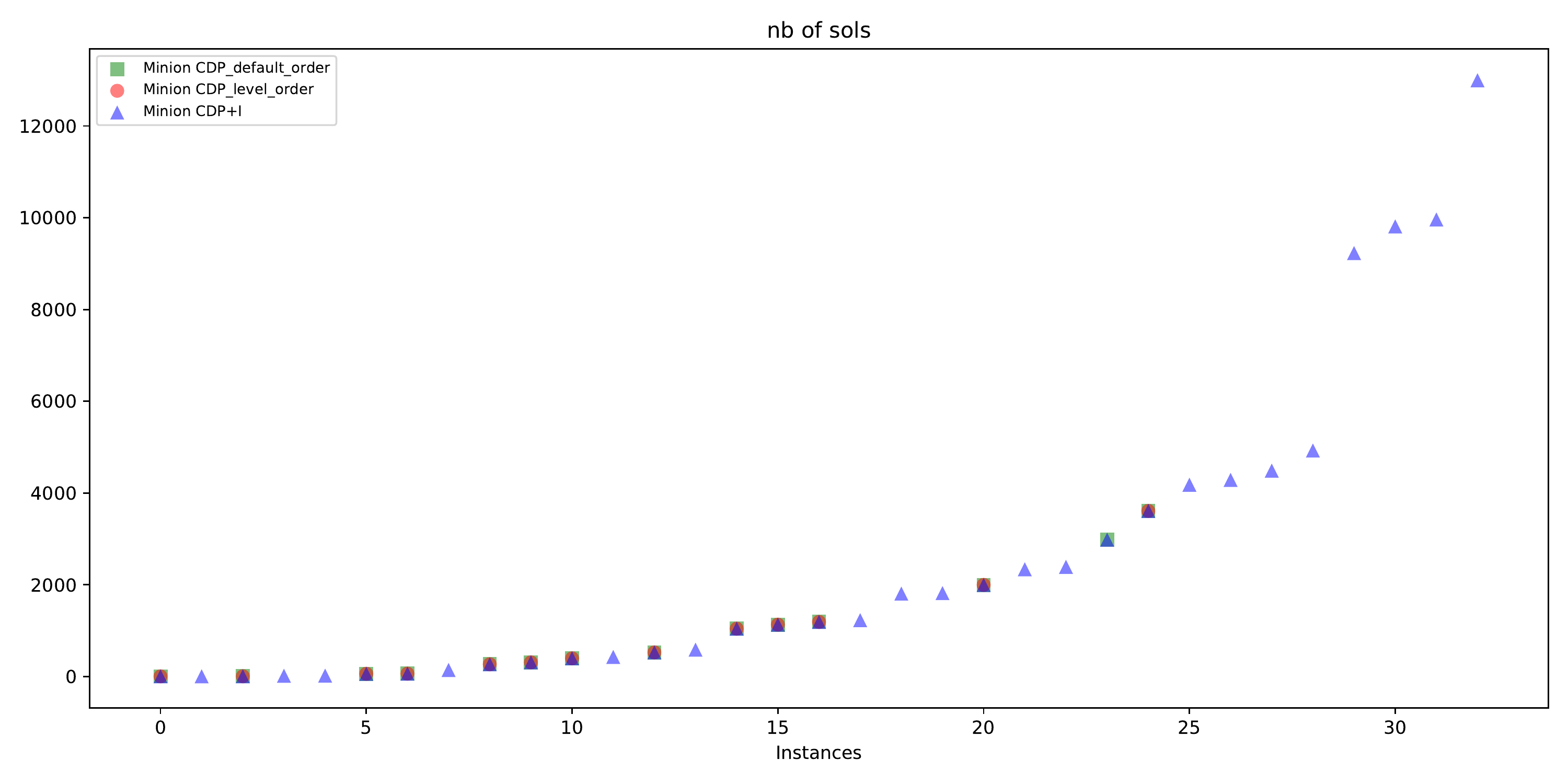}
    \caption{Number of solutions found by three configurations.
    \label{fig:gen_nb}}
\end{figure}

\Cref{fig:gen_nb} shows the number of solutions which have been found by these three configurations. They are sorted according to the number of solutions. Missing bullet-points indicate that that specific instance for that configuration is timed out.

As we can see in this plot, the number of solutions is the same for all the configurations. This is particular to generator itemset mining problem instances using Minion with a static variable ordering. Since generator itemsets built on cardinality in an increasing manner, CDP ones (CDP\_default\_order and CDP\_level\_order) are obliged to find no dominated solutions thanks to the default variable ordering.

\subsubsection{On Solver Time}

\begin{figure}[t]
    \centering
    \includegraphics[width=\textwidth]{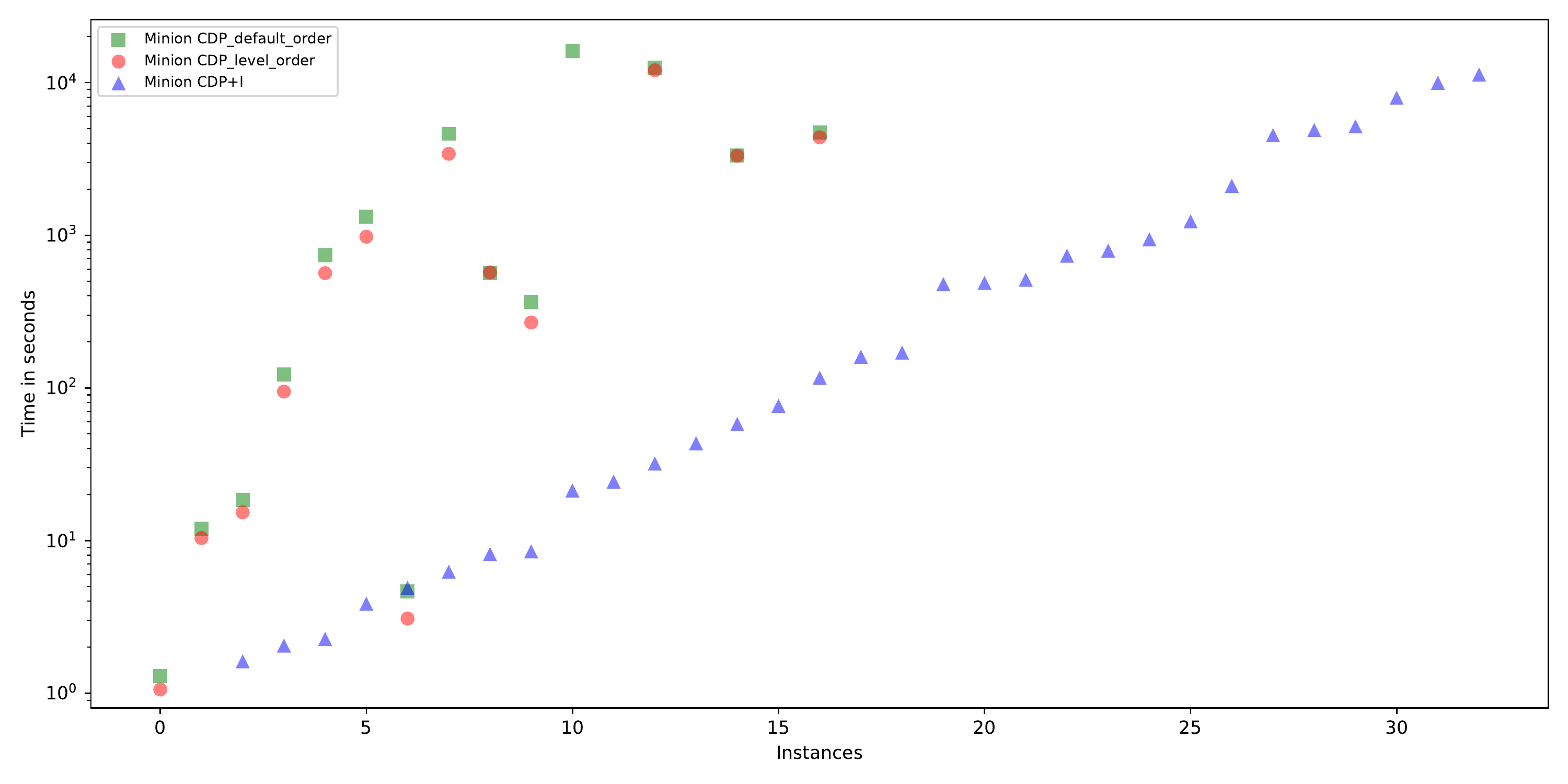}
    \caption{Solver time for all instances, sorted by Minion CDP+I.
    \label{fig:gen_solverlog}}
\end{figure}

\Cref{fig:gen_solverlog} shows the time spent solving each of the instances, using the three configurations, sorted by time taken by Minion CDP+I. We see that in general CDP+I configurations are significantly better than CDP configurations for most instances. For a small number of instances (where there are a small number of solutions to be found.

As we can see, Minion CDP\_level\_order (in red) performs better than the CDP\_default\_order (in green). This is of course expected. Moreover, since CDP+I (in blue) has the same exact search order as CDP\_level\_order, it is fair to compare these two in detail. Even though, this plot show a good overview of this comparison, we can look them side by side to see the affect of the incomparability directly.

\begin{figure}
   \centering
\begin{subfigure}[t]{.45\linewidth}
    \centering
    \includegraphics[width=\textwidth]{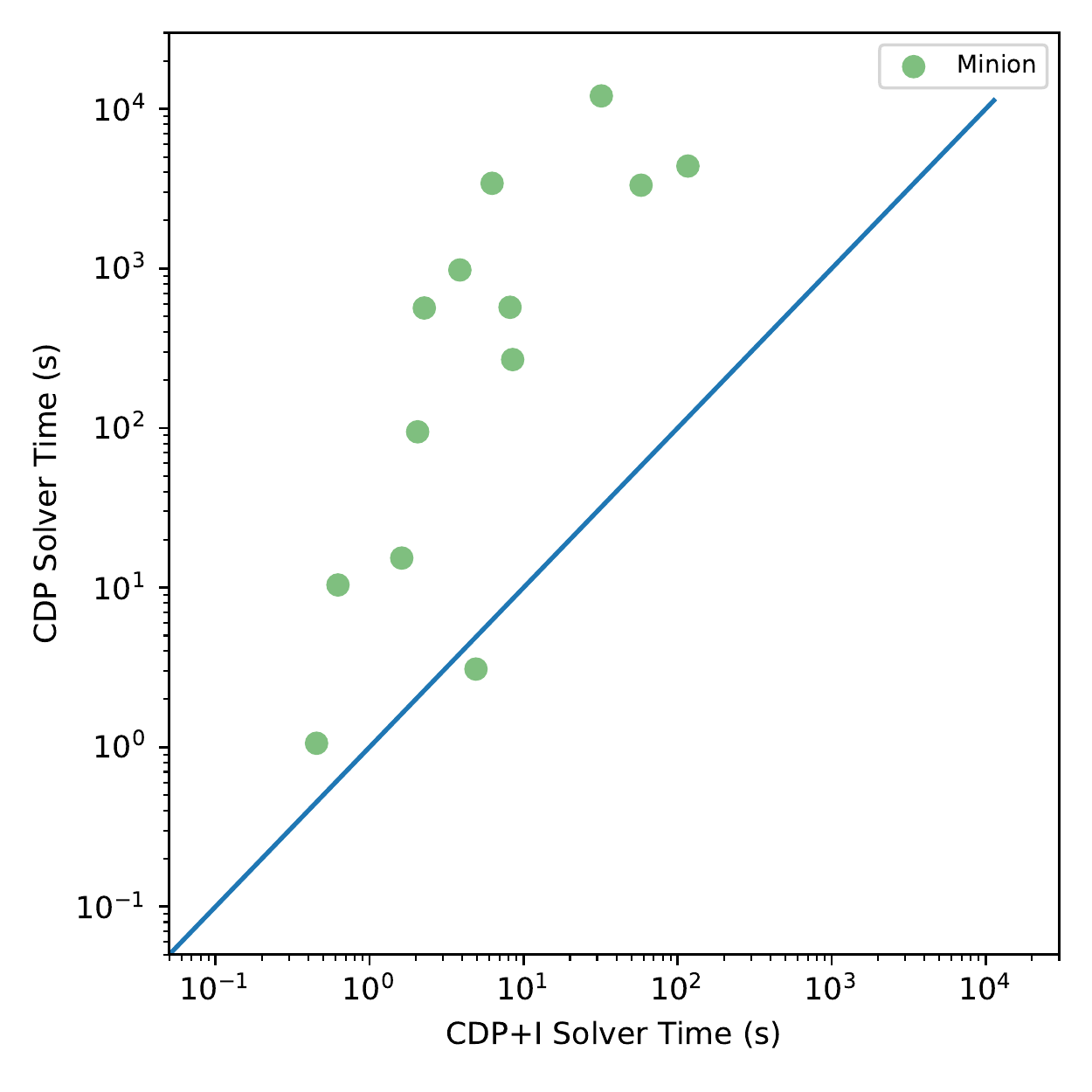}
    \caption{Comparing CDP\_level\_order with CDP+I on Minion.
    \label{fig:comp1} }
\end{subfigure}
\hfill    
\begin{subfigure}[t]{.45\linewidth}
    \centering
    \includegraphics[width=\textwidth]{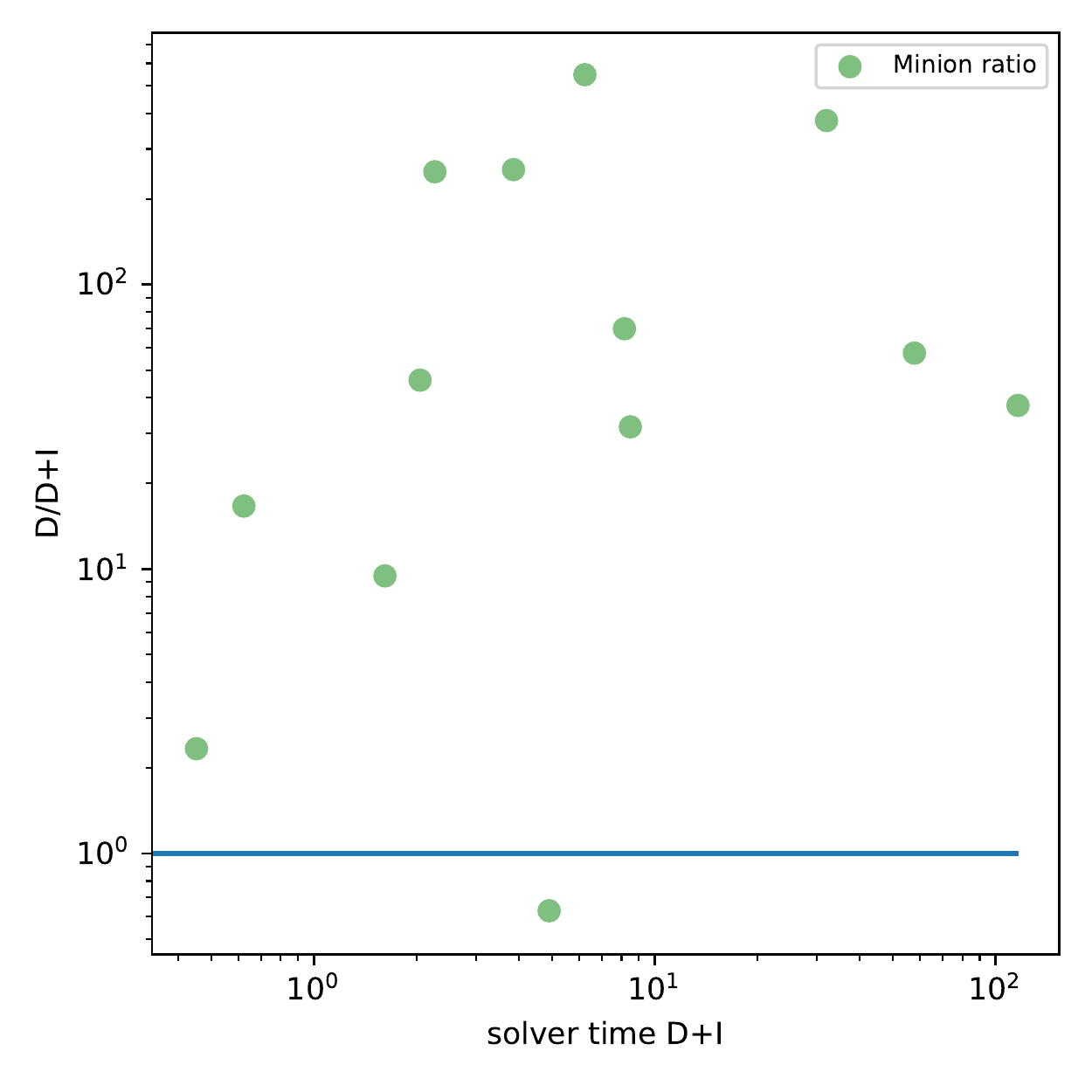}
    \caption{Reduction ration when comparing CDP\_level\_order with CDP+I on Minion.
    \label{fig:comp2}}
\end{subfigure}
\caption{Comparing solver times between the CDP and CDP+I configurations.}
\end{figure}

\Cref{fig:comp1} presents a comparison plot between CDP\_level\_order and CDP+I for \minion. A point above the diagonal line means CDP+I performs betterCDP\_level\_order. In a majority of the instances CDP+I performs better. 

\Cref{fig:comp2} also presents a comparison plot between CDP\_level\_order and CDP+I. But this plot shows the reduction on time by ration on CDP\_level\_order/CDP+I. As we can see, the reduction can be more than 100 fold.

In both of these time comparison plots, there is a losing instance for CDP+I. The reason comes from the number of solver calls.

\subsubsection{On Number of Solver Calls}

\begin{figure}[h!]
    \centering
    \includegraphics[width=\textwidth]{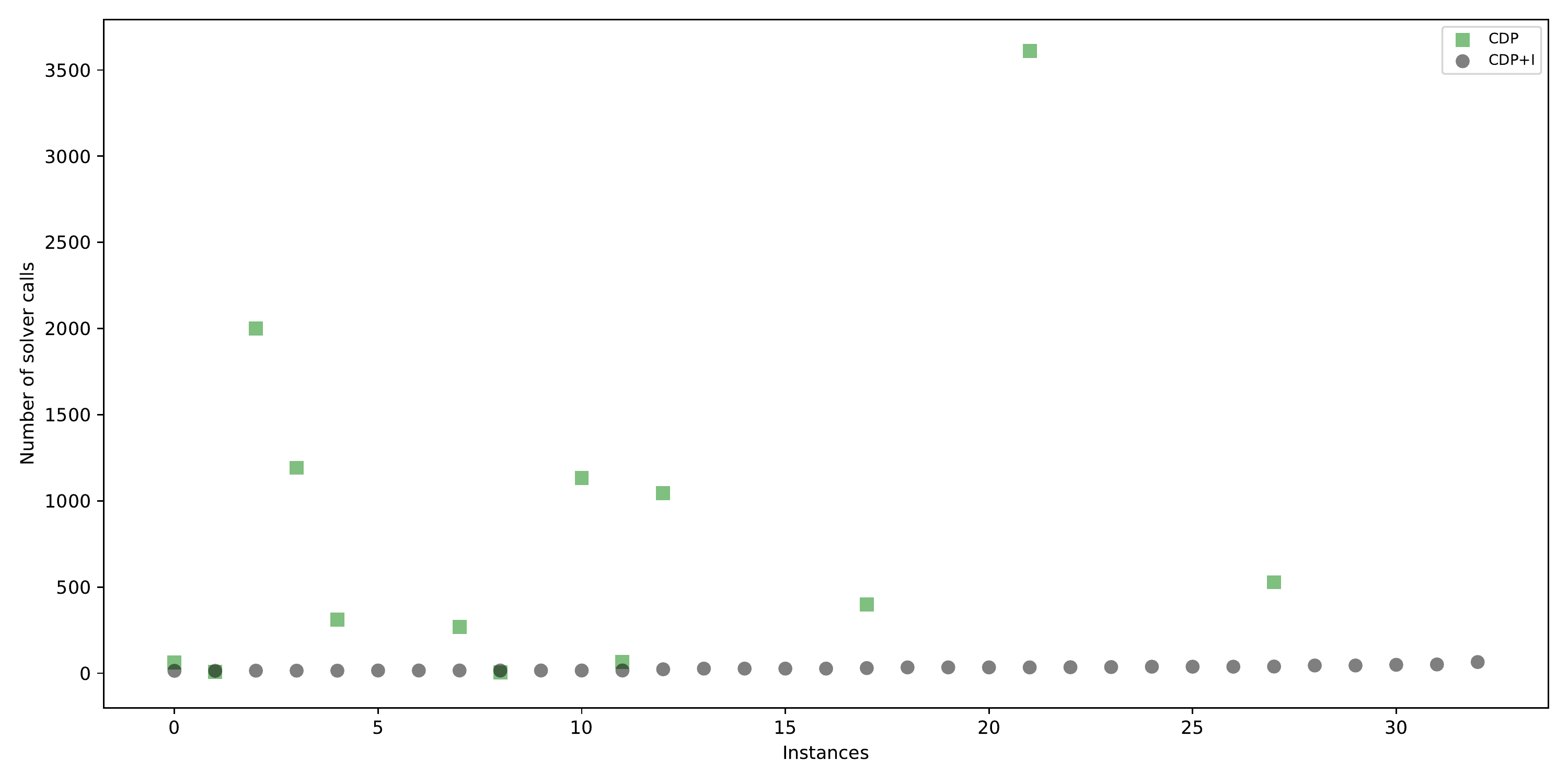}
    \caption{Number of solver calls for the comparison of CDP vs CDP+I \label{fig:nbSolverCalls}}
\end{figure}

\Cref{fig:nbSolverCalls} presents the number of solver calls made by CDP+I and the two CDP configurations. The number of solver calls is identical for CDP configurations since it is one more than the number of solutions. CDP+I makes significantly fewer solver calls, except for a couple of instances where there is a small number of solutions ($< 25$) and a comparatively larger number of levels. 

\subsection{CDP+I's best and worst performances}

When we can look at the instances where CDP+I outperforms CDP, it is due to reducing the number of solver calls drastically. We can see this one particular example in \Cref{tab:hep50}. The number of solver calls is 10 times less and this affects the solver total time drastically. Also, the total blocking clauses used in each solver search is averages out really less as well.

\begin{table}[h!]
\begin{tabular}{|l|r|r|r|r|r|}
\hline
Config & Nb of Sols & Total Solver Time & Total Blocking Clauses & Solver Calls \\ \hline
minion CDP+I         & 3610 & 31.95 & 108546  & 34   \\
minion CDP\_default\_order    & 3610 & 12510.53 & 6517855 & 3611 \\
minion CDP\_level\_order & 3610 & 12059.16 & 6517855 & 3611 \\ \hline
\end{tabular}
\caption{Example of Hepatitis dataset on  50\% frequency \label{tab:hep50}}
\end{table}

If we look at that one single losing instance \Cref{tab:vote40}, even though the solver time is really close, when we look at the ration it is significant. The difference comes from +10 solver calls happened in CDP+I compared to CDP.

\begin{table}[h!]
\begin{tabular}{|l|r|r|r|r|}
\hline
Config & Nb of Sols & Total Solver Time & Total Blocking Clauses & Solver Calls \\ \hline
Minion CDP+I        & 5 & 4.90 & 63 & 16 \\ \hline
Minion CDP\_default\_order          & 5 & 4.64 & 15 & 6  \\ \hline
Minion CDP\_level\_order & 5 & 3.08  & 15 & 6  \\ \hline
\end{tabular}
\caption{Example of Vote dataset on 40\% frequency \label{tab:vote40}}
\end{table}

\section{Conclusion}

In this paper we extended the high-level problem specification language \essence{} to support dominance programming features. In addition, we defined and implement an enhancement to standard CDP in the form of an incomparability function statement. Equipped with CDP+I capabilities, \essence{} becomes a particularly suitable language for specifying and solving constraint-based itemset mining problems that both contain problem specific side constraints and constraints among solutions. We explained our architecture for CDP+I and present preliminary computational experiments on the Generator Itemset Mining problem to show the efficacy of this approach. We showed that by adding an explicit incomparability function to a CDP model, one can achieve significant performance gains and produce a drastically reduced number of dominated solutions.


Future work includes the application of CDP+I to a wider range of problem classes, both in data mining and beyond. We believe multi-objective optimisation problems will be a natural next application area for CDP+I.




\bibliographystyle{splncs04}
\bibliography{ref}

\begin{thebibliography}{10}
\providecommand{\url}[1]{\texttt{#1}}
\providecommand{\urlprefix}{URL }
\providecommand{\doi}[1]{https://doi.org/#1}

\bibitem{agrawal1994fast}
Agrawal, R., Srikant, R., et~al.: Fast algorithms for mining association rules.
  In: Proc. 20th int. conf. very large data bases, VLDB. vol.~1215, pp.
  487--499 (1994)

\bibitem{akgun2013automated}
Akgun, O., Frisch, A.M., Gent, I.P., Hussain, B.S., Jefferson, C., Kotthoff,
  L., Miguel, I., Nightingale, P.: Automated symmetry breaking and model
  selection in conjure. In: International Conference on Principles and Practice
  of Constraint Programming. pp. 107--116. Springer (2013)

\bibitem{akgun2014breaking}
Akgun, O., Gent, I.P., Jefferson, C., Miguel, I., Nightingale, P.: Breaking
  conditional symmetry in automated constraint modelling with conjure. In:
  ECAI. pp.~3--8 (2014)

\bibitem{akgun2011extensible}
Akgun, O., Miguel, I., Jefferson, C., Frisch, A.M., Hnich, B.: Extensible
  automated constraint modelling. In: Twenty-Fifth AAAI Conference on
  Artificial Intelligence (2011)

\bibitem{bonchi2003examiner}
Bonchi, F., Giannotti, F., Mazzanti, A., Pedreschi, D.: Examiner: Optimized
  level-wise frequent pattern mining with monotone constraints. In: Third IEEE
  International Conference on Data Mining. pp. 11--18. IEEE (2003)

\bibitem{bonchi2004closed}
Bonchi, F., Lucchese, C.: On closed constrained frequent pattern mining. In:
  Fourth IEEE International Conference on Data Mining (ICDM'04). pp. 35--42.
  IEEE (2004)

\bibitem{bonchi2007extending}
Bonchi, F., Lucchese, C.: Extending the state-of-the-art of constraint-based
  pattern discovery. Data \& Knowledge Engineering  \textbf{60}(2),  377--399
  (2007)

\bibitem{boulicaut2001mining}
Boulicaut, J.F., Jeudy, B.: Mining free itemsets under constraints. In:
  Proceedings 2001 International Database Engineering and Applications
  Symposium. pp. 322--329. IEEE (2001)

\bibitem{boulicaut2000approximation}
Boulicaut, J.F., Bykowski, A., Rigotti, C.: Approximation of frequency queries
  by means of free-sets. In: European Conference on Principles of Data Mining
  and Knowledge Discovery. pp. 75--85. Springer (2000)

\bibitem{chan2003mining}
Chan, R., Yang, Q., Shen, Y.D.: Mining high utility itemsets. In: Third IEEE
  international conference on data mining. pp. 19--26. IEEE (2003)

\bibitem{de2008constraint}
De~Raedt, L., Guns, T., Nijssen, S.: Constraint programming for itemset mining.
  In: Proceedings of the 14th ACM SIGKDD international conference on Knowledge
  discovery and data mining. pp. 204--212. ACM (2008)

\bibitem{frisch2008ssence}
Frisch, A.M., Harvey, W., Jefferson, C., Mart{\'\i}nez-Hern{\'a}ndez, B.,
  Miguel, I.: E ssence: A constraint language for specifying combinatorial
  problems. Constraints  \textbf{13}(3),  268--306 (2008)

\bibitem{guns2017miningzinc}
Guns, T., Dries, A., Nijssen, S., Tack, G., De~Raedt, L.: Miningzinc: A
  declarative framework for constraint-based mining. Artificial Intelligence
  \textbf{244},  6--29 (2017)

\bibitem{guns2013miningzinc}
Guns, T., Dries, A., Tack, G., Nijssen, S., De~Raedt, L.: Miningzinc: A
  modeling language for constraint-based mining. In: Twenty-Third International
  Joint Conference on Artificial Intelligence (2013)

\bibitem{guns2018solution}
Guns, T., Stuckey, P.J., Tack, G.: Solution dominance over constraint
  satisfaction problems. arXiv preprint arXiv:1812.09207  (2018)

\bibitem{han2004mining}
Han, J., Pei, J., Yin, Y., Mao, R.: Mining frequent patterns without candidate
  generation: A frequent-pattern tree approach. Data mining and knowledge
  discovery  \textbf{8}(1),  53--87 (2004)

\bibitem{koccak2018closed}
Ko{\c{c}}ak, G., Akg{\"u}n, {\"O}., Miguel, I., Nightingale, P.: Closed
  frequent itemset mining with arbitrary side constraints. In: 2018 IEEE
  International Conference on Data Mining Workshops (ICDMW). pp. 1224--1232.
  IEEE (2018)

\bibitem{kryszkiewicz1998representative}
Kryszkiewicz, M.: Representative association rules and minimum condition
  maximum consequence association rules. In: European Symposium on Principles
  of Data Mining and Knowledge Discovery. pp. 361--369. Springer (1998)

\bibitem{negrevergne2013dominance}
Negrevergne, B., Dries, A., Guns, T., Nijssen, S.: Dominance programming for
  itemset mining. In: 2013 IEEE 13th International Conference on Data Mining.
  pp. 557--566. IEEE (2013)

\bibitem{nethercote2007minizinc}
Nethercote, N., Stuckey, P.J., Becket, R., Brand, S., Duck, G.J., Tack, G.:
  Minizinc: Towards a standard cp modelling language. In: International
  Conference on Principles and Practice of Constraint Programming. pp.
  529--543. Springer (2007)

\bibitem{nightingale2014automatically}
Nightingale, P., Akg{\"u}n, {\"O}., Gent, I.P., Jefferson, C., Miguel, I.:
  Automatically improving constraint models in savile row through
  associative-commutative common subexpression elimination. In: International
  Conference on Principles and Practice of Constraint Programming. pp.
  590--605. Springer (2014)

\bibitem{nightingale2017automatically}
Nightingale, P., Akg{\"u}n, {\"O}., Gent, I.P., Jefferson, C., Miguel, I.,
  Spracklen, P.: Automatically improving constraint models in savile row.
  Artificial Intelligence  \textbf{251},  35--61 (2017)

\bibitem{pasquier1999discovering}
Pasquier, N., Bastide, Y., Taouil, R., Lakhal, L.: Discovering frequent closed
  itemsets for association rules. In: International Conference on Database
  Theory. pp. 398--416. Springer (1999)

\bibitem{soulet2014efficiently}
Soulet, A., Rioult, F.: Efficiently depth-first minimal pattern mining. In:
  Pacific-Asia Conference on Knowledge Discovery and Data Mining. pp. 28--39.
  Springer (2014)

\bibitem{szathmary2007towards}
Szathmary, L., Napoli, A., Valtchev, P.: Towards rare itemset mining. In: 19th
  IEEE International Conference on Tools with Artificial Intelligence (ICTAI
  2007). vol.~1, pp. 305--312. IEEE (2007)

\bibitem{zaki2000scalable}
Zaki, M.J.: Scalable algorithms for association mining. IEEE transactions on
  knowledge and data engineering  \textbf{12}(3),  372--390 (2000)

\end{thebibliography}

\end{document}